\pdfoutput=1

\documentclass[11pt]{article}

\usepackage[preprint]{acl}

\usepackage{times}
\usepackage{latexsym}
\usepackage{url}
\usepackage{hyperref}
\usepackage{tabularx}
\usepackage{multirow}
\usepackage{enumitem}
\usepackage{adjustbox}
\usepackage{tcolorbox}
\usepackage{booktabs} 
\usepackage{amsfonts}
\usepackage[T1]{fontenc}
\usepackage[utf8]{inputenc}
\usepackage{microtype}
\usepackage{inconsolata}
\usepackage{graphicx}
\usepackage{xspace}
\usepackage{arydshln}
\usepackage{amsmath}

\setlist{leftmargin=*,nosep}

\title{MedAidDialog: A Multilingual Multi-Turn Medical Dialogue Dataset for Accessible Healthcare}

\author{{Shubham Kumar Nigam}$^{1*\dagger}$ \quad Suparnojit Sarkar$^{2*}$ \quad Piyush Patel$^{3*}$\\
$^{1}$ University of Birmingham, Dubai, United Arab Emirates \\
$^{2}$ Heritage Institute of Technology, Kolkata, India\\
$^{3}$ Madan Mohan Malaviya University of Technology, India\\
\texttt{\{shubhamkumarnigam, suparnojit2026, ppiyush0005\}@gmail.com}
}

\date{}

\begin{document}
\maketitle

\renewcommand{\thefootnote}{$*$}
\footnotetext{These authors contributed equally to this work}
\renewcommand{\thefootnote}{$\dagger$}
\footnotetext{Corresponding author}
\renewcommand{\thefootnote}{\arabic{footnote}}

\begin{abstract}

Conversational artificial intelligence has the potential to assist users in preliminary medical consultations, particularly in settings where access to healthcare professionals is limited. However, many existing medical dialogue systems operate in a single-turn question--answering paradigm or rely on template-based datasets, limiting conversational realism and multilingual applicability. In this work, we introduce \texttt{MedAidDialog}, a multilingual multi-turn medical dialogue dataset designed to simulate realistic physician--patient consultations. The dataset extends the \texttt{MDDial} corpus by generating synthetic consultations using large language models and further expands them into a parallel multilingual corpus covering seven languages: English, Hindi, Telugu, Tamil, Bengali, Marathi, and Arabic. 
Building on this dataset, we develop \texttt{MedAidLM}, a conversational medical model trained using parameter-efficient fine-tuning on quantized small language models, enabling deployment without high-end computational infrastructure. Our framework additionally incorporates optional patient pre-context information (e.g., age, gender, allergies) to personalize the consultation process. Experimental results demonstrate that the proposed system can effectively perform symptom elicitation through multi-turn dialogue and generate diagnostic recommendations. We further conduct medical expert evaluation to assess the plausibility and coherence of the generated consultations. 

\end{abstract}

\section{Introduction}
Conversational artificial intelligence has recently demonstrated strong potential for assisting users in healthcare settings, particularly for preliminary symptom assessment and medical guidance. Large language models (LLMs) have shown impressive capabilities in natural language understanding and dialogue generation, enabling systems to interact with patients in a conversational manner \citep{tu2024towards}. However, many existing models primarily operate in a \textit{single-turn} question–answering paradigm, where users provide all relevant information in a single prompt. In real clinical practice, physicians rarely rely on such interactions; instead, diagnosis typically emerges through a sequence of questions that progressively refine the patient's symptoms.

Furthermore, most conversational medical AI systems are trained on datasets that are either template-based or limited to a single language. While datasets such as \texttt{MDDial} \citep{macherla2023mddial} provide an important step toward multi-turn diagnostic dialogue, template-driven generation often constrains linguistic diversity and conversational realism. In addition, the lack of multilingual dialogue resources limits the applicability of such systems in low-resource environments, where patients may not communicate in English.

\begin{figure}[t]
\centering
\includegraphics[width=\linewidth]{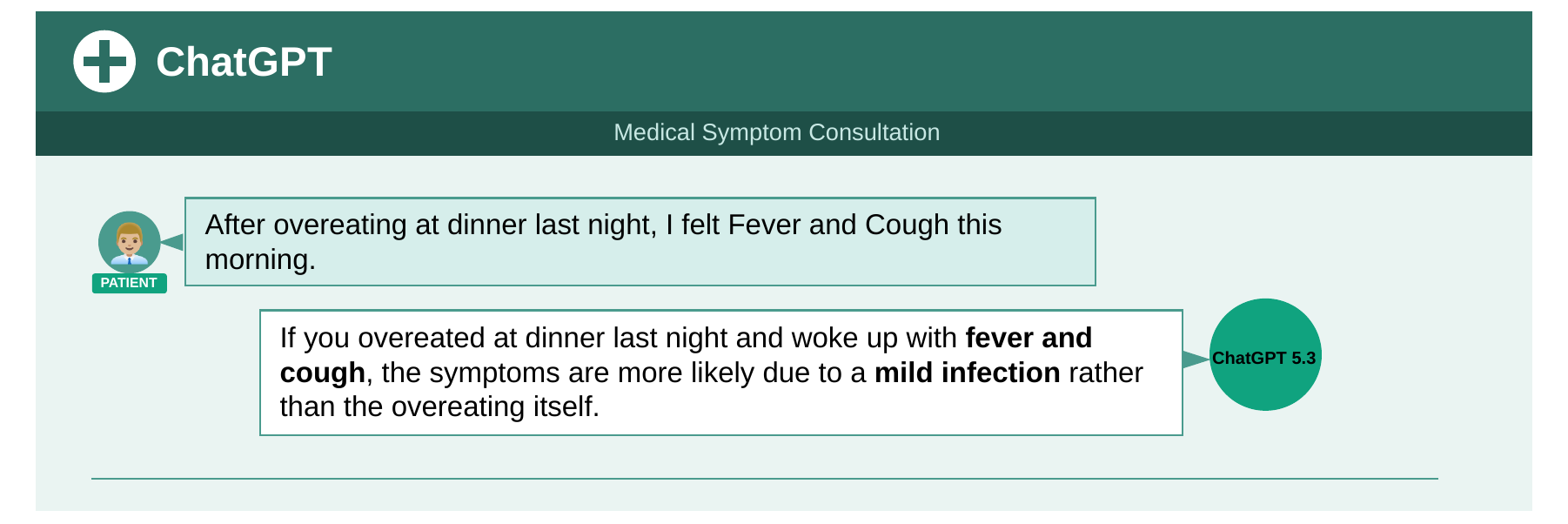}
\caption{Example response from a general-purpose LLM (ChatGPT 5.3). The model produces a single explanatory response without collecting additional symptoms or conducting follow-up questioning.}
\label{fig:gpt_example}
\end{figure}

Another important limitation of many existing systems is the absence of \textit{patient context}. In real consultations, physicians typically begin with basic demographic information such as age, gender, medical history, or allergies before asking symptom-related questions. Without this information, responses generated by general-purpose models may remain generic or overly verbose. Figure~\ref{fig:gpt_example} illustrates this limitation: a general-purpose LLM generates a single explanatory answer without conducting follow-up questioning, whereas our proposed model engages in a multi-turn dialogue to collect additional symptoms before providing a diagnostic recommendation.

To address these limitations, we introduce \texttt{MedAidDialog}, a multilingual multi-turn medical dialogue dataset designed to simulate realistic physician–patient consultations. The dataset extends the \texttt{MDDial} corpus with synthetic dialogues generated using a large language model and further expands the conversations into a parallel multilingual corpus covering seven languages: English, Hindi, Telugu, Tamil, Bengali, Marathi, and Arabic. This multilingual design aims to improve accessibility of conversational healthcare systems for users in rural or linguistically diverse regions.

Building on this dataset, we develop \texttt{MedAidLM}, a fine-tuned conversational medical model trained using parameter-efficient fine-tuning techniques. Unlike large proprietary systems that require extensive computational resources,  is trained using quantized small language models and can therefore be deployed on modest hardware environments. This makes the approach particularly suitable for low-resource healthcare settings where high-end infrastructure may not be available.

\begin{figure}[t]
\centering
\includegraphics[width=\linewidth]{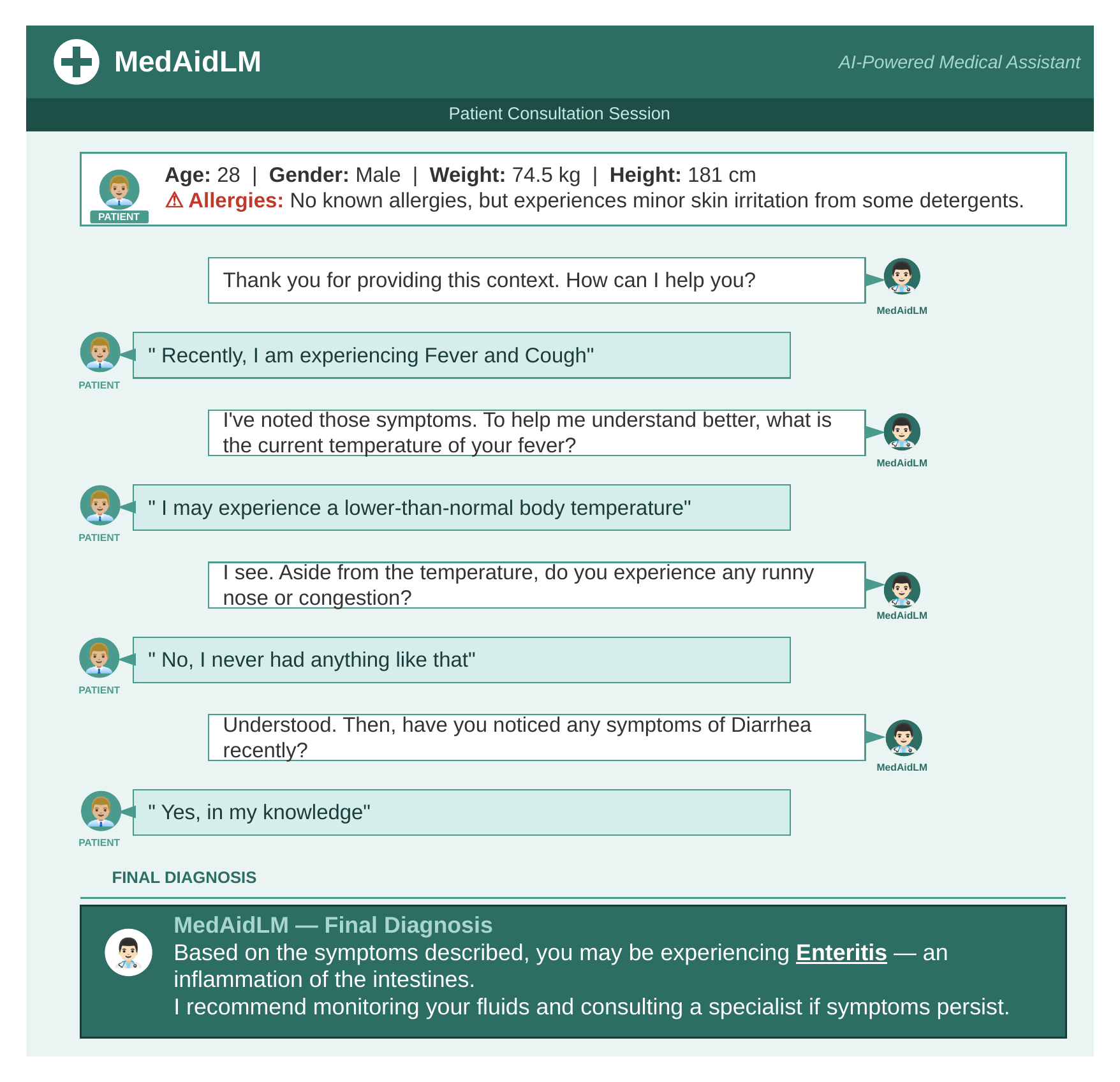}
\caption{Example interaction with \texttt{MedAidLM}. The system first incorporates patient pre-context information (e.g., age, gender, and allergies) and then performs multi-turn dialogue to collect symptoms before producing a diagnostic recommendation.}
\label{fig:medaid_example}
\end{figure}

Figure~\ref{fig:gpt_example} illustrates the behavior of a general-purpose LLM, which generates a single verbose response without engaging in follow-up questioning. In contrast, the proposed  system (Figure~\ref{fig:medaid_example}) utilizes patient pre-context information and performs multi-turn conversational symptom elicitation before producing a diagnosis, more closely resembling a real physician–patient consultation.

To ensure reliability of the generated consultations, we additionally conduct evaluation with medical experts who assess the coherence and plausibility of the model’s responses. This evaluation provides qualitative validation of the system’s ability to simulate realistic clinical dialogue.

To ensure reproducibility and encourage further research, the dataset and model code will be made publicly available soon.

\paragraph{Contributions}

The main contributions of this work are summarized as follows:

\begin{itemize}

\item We introduce a new task of \textit{multilingual multi-turn medical dialogue generation} and construct \texttt{MedAidDialog}, a parallel medical dialogue dataset designed for low-resource multilingual environments.

\item We incorporate \textit{patient pre-context information} (e.g., age, gender, allergies, and demographic attributes) to enable personalized conversational medical assistance.

\item We develop \texttt{MedAidLM}, a parameter-efficient fine-tuned conversational model based on quantized small language models, enabling deployment without high-end computational infrastructure.

\item We perform \textit{medical expert evaluation} to validate the quality and plausibility of the generated diagnostic dialogues.

\end{itemize}

 \section{Related Work}

Prior work on medical dialogue has progressed from structured and task-oriented diagnosis systems toward neural and LLM-based conversational assistants. Early datasets and systems emphasized symptom collection, slot filling, or diagnosis prediction, but often lacked natural multi-turn physician--patient interaction \citep{zeng2020meddialog,liu2022meddg}. More recent resources explicitly target multi-turn medical consultation. For example, MDDial introduces an English differential-diagnosis dialogue dataset, but it is constructed through templates and remains partially scripted \citep{macherla2023mddial}. MedDG and Zhongjing advance multi-turn medical conversation in Chinese, with a focus on entity-aware consultation and improving proactive inquiry using real-world dialogue \citep{liu2022meddg,yang2024zhongjing}. MediTOD further provides an English medical history-taking dataset with detailed annotations, though it is primarily designed for structured task-oriented interaction \citep{saley2024meditod}.

In parallel, medical LLMs such as ChatDoctor, Med-Chat, and related systems have shown that domain-specific fine-tuning substantially improves medical response quality over general-purpose LLMs \citep{li2023chatdoctor,10911671}. However, many such systems are still optimized for single-turn question answering or instruction following, which assumes that patients can provide complete and precise information in one prompt. This differs from real clinical practice, where doctors iteratively ask follow-up questions before giving advice or forming a diagnosis. AMIE frames diagnosis as conversational history-taking and reasoning \citep{tu2024towards}, while DoctorAgent-RL further models multi-turn clinical dialogue as an adaptive decision process with RL \citep{feng2025doctoragent}. Other approaches, such as BianQue, T-Agent, and continuous entity reasoning, explicitly model questioning behavior, medical term flow, or entity transitions across dialogue turns \citep{chen2023bianque,10650649,wang2025continuous}.

Because real clinical conversations are difficult to release due to privacy and governance constraints, several studies have explored synthetic dialogue generation. NoteChat generates patient--physician conversations conditioned on clinical notes \citep{wang2024notechat}, while MDDial uses template-based synthesis from structured diagnostic data \citep{macherla2023mddial}. Such work shows the value of synthetic data for training conversational medical systems, but most existing datasets remain either single-language, template-constrained, or not designed as multilingual parallel corpora.

Multilingual medical dialogue remains especially underexplored. BiMediX is an important step toward bilingual medical conversation in English and Arabic \citep{pieri2024bimedix}, but broader multilingual coverage for low-resource settings is still missing. This limitation is critical for practical deployment, especially in regions where patients may not be comfortable using English and where lightweight models are preferable for accessibility. More broadly, multi-turn dialogue research in NLP has highlighted the importance of context tracking, coherence, reasoning, and safety across turns \citep{li2017dailydialog,cui2020mutual,su2019improving,zhang2021advances,10.1145/3771090,zhou2024speak}. Recent evaluation work in medical dialogue also shows that success should not be measured only by final-answer accuracy, but also by questioning quality, safety, and turn-level clinical relevance \citep{macherla2023mddial,tu2024towards,gong2026meddialogrubrics}.


\section{Task Definition}

We study the problem of multilingual multi-turn medical dialogue generation, where a conversational agent interacts with a patient to collect symptoms and provide preliminary diagnostic guidance. Unlike single-turn medical question answering, this task requires modeling sequential physician--patient interactions where diagnostic reasoning emerges through multiple conversational exchanges.

\subsection{Problem Setup}

A medical consultation dialogue is represented as a sequence of conversational turns between a patient and a doctor $D = \{u_1, u_2, ..., u_T\}$,
where $u_t$ denotes the utterance at turn $t$, and $T$ is the total number of dialogue turns. In our setting, odd turns correspond to patient utterances and even turns correspond to doctor responses. Each dialogue is associated with a diagnostic label $y$ drawn from a disease set $y \in \mathcal{Y}$, where $\mathcal{Y}$ denotes the set of possible diseases considered in the dataset.

Given a dialogue context consisting of the previous turns:

\begin{equation}
C_t = \{u_1, u_2, ..., u_{t-1}\}
\end{equation}

The objective of the model is to generate the next doctor response:

\begin{equation}
u_t = \arg\max_{u} P(u \mid C_t)
\end{equation}

The conversation continues until sufficient information has been collected and a diagnostic recommendation is produced.

\subsection{Multilingual Dialogue Setting}
The dataset supports multilingual dialogue generation across seven languages: English, Hindi, Telugu, Tamil, Bengali, Marathi, and Arabic. The objective is to learn a model that can generate medically coherent responses across languages while maintaining consistent diagnostic reasoning.

\subsection{Patient Context Personalization}

In real clinical consultations, physicians often begin with basic contextual information about the patient before asking symptom-related questions. To better simulate this scenario, our framework allows optional {patient pretext information} to be provided at the start of the dialogue. This information may include \textit{age group, gender, geographic location, known allergies, and pre-existing medical conditions, etc}. This information is appended to the dialogue prefix and incorporated into the model input. Incorporating patient context allows the model to personalize its questioning strategy and diagnostic reasoning, reflecting how clinicians adapt their inquiries based on patient demographics and medical history.





\section{\texttt{MedAidDialog} Dataset}

Multi-turn conversational datasets are essential for training medical dialogue systems that can iteratively collect symptoms and provide diagnostic guidance \citep{macherla2023mddial,tu2024towards}. The MDDial dataset \citep{macherla2023mddial} provides an English differential-diagnosis dialogue corpus derived from structured medical records. However, its template-based generation limits conversational diversity and realism, and it does not support multilingual deployment.

To address these limitations, we construct \texttt{MedAidDialog}, a synthetic multilingual medical dialogue dataset designed to simulate more natural physician--patient consultations while enabling accessibility across multiple languages.

\subsection{Synthetic Dialogue Generation}

To increase conversational diversity beyond template-based dialogues, we generate synthetic medical consultations using the Llama-3.3-70B-Versatile model through the Groq API\footnote{\url{https://groq.com/}}. The model architecture follows the design described in the Llama~3 model card \citep{llama3modelcard}.

The generation pipeline simulates diagnostic consultations involving 12 diseases and 118 symptoms. Each dialogue begins with a randomized patient complaint and proceeds through multiple conversational exchanges in which the physician asks follow-up questions to gather diagnostic evidence. Dialogues typically contain 4--8 conversational turns and conclude with a final diagnosis.

To better approximate real clinical conversations, the generation process introduces variability through non-deterministic patient responses, overlapping symptom descriptions, and incomplete or ambiguous symptom reporting. Using this pipeline, we generated 1,101 synthetic consultations, providing a more diverse training resource compared with template-based dialogue construction. Table~\ref{tab:dataset_statistics} summarizes the statistics of the original MDDial dataset and the synthetic dialogues used to construct \texttt{MedAidDialog}. Compared with the template-driven corpus, the synthetic dataset contains longer dialogues and richer conversational exchanges.
\begin{table}[t]
\centering
\resizebox{\linewidth}{!}{%
\begin{tabular}{lcccccccc}
\toprule
 &
\multicolumn{4}{c}{\textbf{Dialogue Turns}} &
\multicolumn{3}{c}{\textbf{Average Words}} \\

\cmidrule(lr){2-5}
\cmidrule(lr){6-8}
\textbf{Dataset}
&

Avg &
Total &
Min &
Max &
Per &
Patient &
Doctor \\
&
Turns &
Dialogues &
Turns &
Turns &
Dialogue &
Utterance &
Utterance \\

\midrule

MDDial (MD) & 4.9 & 1879 & 1 & 16 & 53.5 & 5.6 & 6.7 \\

Synthetic (SYN) & 6.6 & 1101 & 5 & 11 & 134.5 & 8.8 & 9.6 \\

MD + SYN & 5.7 & 2980 & 1 & 16 & 86.9 & 7.00 & 8.05 \\

MDDial Test & 5.9 & 237 & 1 & 13 & 55.4 & 5.6 & 6.6 \\

\bottomrule
\end{tabular}}
\caption{Statistics of the original MDDial dataset (MD) and the synthetic dialogues used to construct the MedAidDialog corpus. The synthetic dialogues contain more conversational turns and longer utterances, resulting in richer physician–patient interactions.}
\label{tab:dataset_statistics}
\end{table}

\subsection{Multilingual Expansion}

A primary goal of \texttt{MedAidDialog} is to support healthcare accessibility for users in rural or linguistically diverse regions. To this end, we construct a parallel multilingual corpus by translating the English dialogues into six additional languages: \textit{Hindi, Telugu, Tamil, Bengali, Marathi, and Arabic}. Each dialogue therefore has aligned translations across seven languages.
The translation pipeline combines TranslateGemma \citep{finkelstein2026translategemma} and TinyAya \citep{salamanca2026tinyayabridgingscale}, two multilingual models designed for efficient translation and cross-lingual generation. To ensure consistent translation and preservation of medical semantics, we employ a structured prompting strategy. The full translation prompt used in the pipeline is provided in Appendix~\ref{app:prompt_translation}.

\section{Methodology}
\label{sec:methodology}

\begin{figure*}[t]
\centering
\includegraphics[width=\linewidth,height=0.45\textheight]{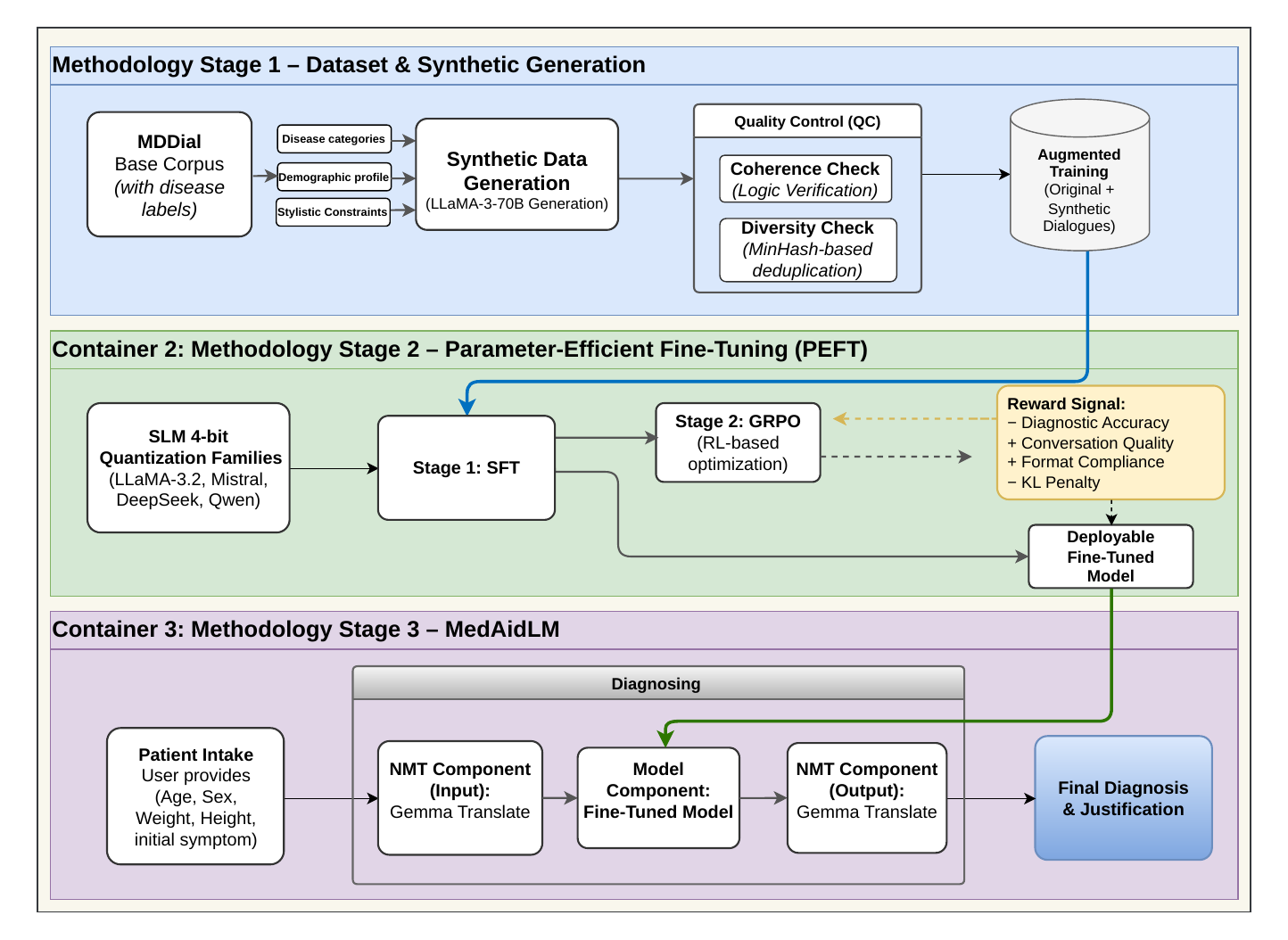}
\caption{
Overview of the proposed framework.
\textbf{Stage 1: Data Augmentation.} The \texttt{MDDial} dataset is expanded with synthetic medical dialogues, followed by coherence and diversity filtering.
\textbf{Stage 2: Model Adaptation.} Compact open-source language models are fine-tuned using parameter-efficient training and LoRA-based SFT. The dotted connection indicates an optional GRPO optimisation stage applied to selected models.
\textbf{Stage 3: Deployment.} The best-performing checkpoint is deployed as \texttt{MedAidLM}, which operates within a multilingual inference loop that incorporates optional patient pre-context and bidirectional translation.
}
\label{fig:architecture_pipeline}
\end{figure*}

Our framework consists of three stages:
(1) synthetic dialogue generation based on \texttt{MDDial},
(2) parameter-efficient fine-tuning of compact open-source language models,
and (3) deployment of the best-performing model in a multilingual conversational system.
Figure~\ref{fig:architecture_pipeline} presents the overall pipeline. 

\subsection{Base Dataset and Synthetic Augmentation}
\label{subsec:synthetic_generation}

We use \texttt{MDDial}~\citep{macherla2023mddial} as the starting point for our data construction pipeline. \texttt{MDDial} is a benchmark corpus for multi-turn medical dialogue in which each conversation is associated with a final disease label. It provides a useful foundation for diagnosis-oriented dialogue modeling, but its template-driven construction limits conversational diversity and does not fully reflect the variability of realistic physician--patient interaction.

To address this limitation, we generate synthetic consultations using Llama-3.3-70B-Versatile via the Groq API.\footnote{\url{https://groq.com/}} The synthetic generation process is conditioned on disease categories from \texttt{MDDial}, demographic profiles, and stylistic constraints so that the generated conversations remain medically plausible while exhibiting richer linguistic variation. The full synthetic generation prompt is included in {Appendix~\ref{app:prompt_syn}}.

Each synthetic consultation is designed to follow a realistic diagnostic flow: the patient presents an initial complaint, the model playing the doctor asks follow-up questions to elicit additional symptoms, and the conversation ends with a diagnosis-oriented response. We target dialogues of 4--8 turns so that the synthetic corpus remains compatible with the interaction style of \texttt{MDDial} while supporting greater diversity in phrasing and symptom progression.

\paragraph{Quality Control.}
To improve the quality of the generated corpus, we apply two filtering stages. First, we perform a \textit{coherence check} to verify logical consistency between symptom descriptions and the final diagnosis. Second, we apply a \textit{diversity check} based on MinHash-style near-duplicate removal to reduce repetitive generations. The resulting synthetic dialogues are merged with the original \texttt{MDDial} training split to form the augmented training corpus, denoted by $\mathcal{D}_{\text{train}}$.

\subsection{Dialogue Formatting}
\label{subsec:dialogue_formatting}

Before training, all dialogues are converted into a unified multi-turn instruction format. Specifically, we transform each consultation into a ShareGPT-style conversation in which patient utterances are mapped to \texttt{human} turns and doctor utterances are mapped to \texttt{gpt} turns. A system message defines the diagnostic consultation setting, and the final assistant turn contains the diagnosis-oriented output. This representation is convenient for instruction tuning and preserves the sequential nature of symptom elicitation. The exact formatting prompt is provided in {Appendix~\ref{app:prompt_sharegpt}}.

\subsection{Parameter-Efficient Fine-Tuning}
\label{subsec:peft_training}

\paragraph{Model Families.}
We fine-tune multiple compact open-source model families in order to study the feasibility of low-resource deployment. Our experiments focus on SLMs, including {Llama-3.2-3B-Instruct}~\citep{grattafiori2024llama}, {Mistral-7B-Instruct}~\citep{jiang20236g}, {DeepSeek-R1-Distill-Qwen-1.5B}~\citep{deepseekai2025deepseekr1incentivizingreasoningcapability}, and {Qwen3-4B}~\citep{qwen3technicalreport}. All models are loaded in 4-bit NF4 quantized format to reduce memory usage and enable training on commodity GPUs.


\paragraph{LoRA Setup.}
We adopt Low-Rank Adaptation (LoRA)~\citep{hu2022lora} for parameter-efficient fine-tuning. LoRA adapters are inserted into the attention projection layers of each transformer block, enabling efficient adaptation while keeping the number of trainable parameters small. Detailed hyperparameters and configuration settings are provided in Appendix~\ref{app:lora}.

\paragraph{Stage 1: Supervised Fine-Tuning.}
In the first training stage, each model is fine-tuned on $\mathcal{D}_{\text{train}}$ using standard next-token prediction. We train for three epochs using AdamW~\citep{loshchilov2017decoupled} with a cosine learning-rate schedule. The dialogues are formatted so that the model learns to ask symptom-focused follow-up questions and delay disease prediction until enough information has been collected.


\paragraph{Optional RL Optimisation.}
Starting from the supervised checkpoint, we optionally apply Group Relative Policy Optimisation (GRPO)~\citep{shao2024deepseekmath} to further refine dialogue behaviour. The reward signal combines diagnostic correctness, conversational quality, and format compliance. Since this optimisation step is optional and not used in all model variants, additional implementation details are provided in Appendix~\ref{app:grpo}.

\subsection{Patient Pre-Context and Personalisation}
\label{subsec:precontext}

A key component of our framework is the use of optional \textit{patient pre-context} before the dialogue begins. This pre-context may include demographic or clinically useful attributes such as age, gender, height, weight, allergies, and other basic history fields. We prepend this information to the conversation as a structured consultation profile, allowing the model to condition its questioning strategy on essential patient characteristics.

This design more closely matches real consultation settings, where physicians often begin with basic contextual information before exploring symptoms in detail. It also enables more personalized follow-up questions, especially in cases where age, sex, or allergy information may influence diagnostic reasoning.

\subsection{Multilingual Inference Pipeline}
\label{subsec:multilingual_inference}

The best-performing fine-tuned checkpoint is deployed as \texttt{MedAidLM}, the dialogue engine in our multilingual consultation system. Since the fine-tuned model operates in English, we wrap it with a bidirectional translation layer so that patients can interact in their preferred language.

At inference time, the user input in language $\ell$ is first translated into English, then passed to \texttt{MedAidLM} together with the patient pre-context and dialogue history. The model generates the next English response, which is then translated back into the user language before being displayed. This process continues turn by turn until the model emits a dedicated \texttt{[PREDICT]} marker, after which the final diagnosis and justification are returned.

For the translation layer, we evaluate {TranslateGemma}~\citep{finkelstein2026translategemma} and {TinyAya}~\citep{salamanca2026tinyayabridgingscale}. The final system prompt used for translation is shown in {Appendix~\ref{app:prompt_translation}}. This translation-augmented loop enables multilingual use while preserving a single English-centered dialogue model.

\subsection{System Summary}
\label{subsec:system_summary}

The resulting system combines data augmentation, compact model adaptation, and multilingual inference into a single deployable pipeline. Synthetic augmentation improves conversational diversity, LoRA-based tuning enables efficient adaptation of compact models, and the translation wrapper allows the system to serve users across multiple low-resource languages without requiring a separate dialogue model per language.

\section{Evaluation Metrics}
\label{sec:evaluation}

\begin{table}[t]
\centering
\resizebox{\linewidth}{!}{
\begin{tabular}{lccc}
\toprule
\textbf{Model Family} & \textbf{Dataset} & \textbf{Method} & \textbf{Accuracy} \\
\midrule
Mistral-7B-Instruct & MedAidDialog (MD+SYN) & SFT & 88.09\% \\
\textbf{LLaMA 3.2 3B (MedAidLM)} & \textbf{MedAidDialog (MD+SYN)} & \textbf{SFT} & \textbf{90.21\%} \\
Qwen3-4B & MedAidDialog (MD+SYN) & SFT & 80.00\% \\
DeepSeek-R1-Distill-Qwen-1.5B & MedAidDialog (MD+SYN) & SFT & 40.00\% \\
\bottomrule
\end{tabular}}
\caption{Results on the  \texttt{MedAidDialog} dataset.}
\label{tab:main_results}
\end{table}


Evaluating conversational medical systems is challenging because a correct diagnosis alone does not guarantee a safe or clinically meaningful interaction. Therefore, we adopt a two-stage evaluation strategy consisting of (i) automatic evaluation based on diagnostic accuracy and (ii) human expert evaluation focusing on clinical reliability and conversational quality.

\subsection{Automatic Evaluation}
\label{subsec:automatic_eval}

For automatic evaluation, we compute the diagnostic accuracy of the model. 
Specifically, we compare the final diagnosis predicted by the model with the gold disease label provided in the dataset. Although accuracy provides a straightforward measure of diagnostic correctness, it does not capture other critical aspects of conversational medical systems such as safety, reasoning quality, or conversational coherence. Therefore, we complement automatic evaluation with human expert assessment.

\subsection{Expert Evaluation}
\label{subsec:expert_eval}

To assess the clinical reliability of the generated conversations, we conduct a human evaluation with three medical experts. All evaluators are qualified medical practitioners holding an \textit{MBBS degree} and are currently pursuing postgraduate medical training at a reputed medical institute. Their medical background enables them to critically evaluate the plausibility, safety, and clinical reasoning of the generated dialogues.

Each expert independently reviewed a subset of randomly sampled dialogues produced by the system. The evaluation focuses on multiple aspects of conversational medical assistance, including safety, symptom understanding, contextual reasoning, diagnostic plausibility, and conversational quality. Most criteria are scored on a \textit{Likert scale from 1 (Very Poor) to 5 (Excellent)}, while medical safety is evaluated as a binary \textit{pass/fail} metric.
Table~\ref{tab:expert_eval_metrics} in Appendix summarizes the evaluation criteria used in the expert assessment.


\section{Results and Analysis}
\label{sec:results}




Table~\ref{tab:main_results} presents the main automatic evaluation results, reporting only the best-performing configuration for each model family trained on the final \texttt{MedAidDialog} corpus. Among all evaluated compact models, LLaMA3.2-3B achieves the highest diagnostic accuracy of \textbf{90.21\%}, and we designate this final model as \texttt{MedAidLM}. Mistral-7B-Instruct also performs strongly with 88.09\% accuracy, whereas Qwen3-4B reaches 80.00\%. DeepSeek-R1-Distill-Qwen-1.5B performs substantially worse, suggesting that very small distilled reasoning models may be less suitable for this dialogue-driven medical prediction setting.
These results indicate that compact open-source models can achieve strong diagnostic performance when trained on the augmented \texttt{MedAidDialog} corpus, even without relying on large proprietary systems.

\begin{table}[t]
\centering
\resizebox{\linewidth}{!}{
\begin{tabular}{lccccc}
\toprule
\textbf{Model} & \textbf{Dataset} & \textbf{Avg. Turns} & \textbf{Dialogs} & \textbf{Method} & \textbf{Accuracy} \\
\midrule
Mistral-7B-Instruct & MD & 4.90 & 1879 & SFT & 18.72\% \\
Mistral-7B-Instruct & SYN & 7.28 & 1101 & SFT & 61.28\% \\
Mistral-7B-Instruct & MD+SYN & 5.78 & 2980 & SFT & 80.85\% \\
Mistral-7B-Instruct & MD+SYN & 5.78 & 2980 & SFT+GRPO & 77.87\% \\
\midrule
LLaMA 3.2 3B & MD & 4.90 & 1879 & SFT & 75.74\% \\
LLaMA 3.2 3B & SYN & 7.28 & 1101 & SFT & 71.97\% \\
LLaMA 3.2 3B & MD+SYN & 5.78 & 2980 & SFT & 77.87\% \\
LLaMA 3.2 3B & MD+SYN & 5.78 & 2980 & SFT+GRPO & 43.83\% \\
\midrule
Qwen3-4B & MD+SYN & 5.78 & 2980 & SFT & 80.00\% \\
DeepSeek-R1 & MD+SYN & 5.78 & 2980 & SFT & 40.00\% \\
\bottomrule
\end{tabular}}
\caption{Ablation study over training data composition and optimisation strategy. These experiments correspond to shorter training runs (100 steps), used to analyze the effect of original data (MD), synthetic data (SYN), and the combined \texttt{MedAidDialog} corpus (MD+SYN).}
\label{tab:ablation_results}
\end{table}

\subsection{Ablation Study}
Table~\ref{tab:ablation_results} presents the ablation study analyzing the impact of dataset composition and training strategy. We observe that training on either the original \texttt{MDDial} dataset or the synthetic corpus alone leads to weaker performance compared to training on the combined dataset. This indicates that the two sources provide complementary supervision signals: the original data captures realistic clinical dialogue patterns, while the synthetic augmentation increases linguistic diversity and symptom coverage.
Overall, the results show that synthetic augmentation is most effective when used to complement the original diagnosis-oriented dialogues rather than replacing them.
We also observe that applying GRPO-based optimisation does not consistently outperform supervised fine-tuning alone. This suggests that the supervised signal provided by the combined multi-turn dialogue corpus is already sufficiently strong, and additional reward-based optimisation may introduce training instability without providing consistent benefits.

\subsection{Expert Evaluation and IAA Scores}
As shown in Table~\ref{tab:expert_eval}, \texttt{MedAidLM} achieves a \textbf{95.3\%} medical safety pass rate, indicating that unsafe advice is rare in the sampled dialogues.
The model also obtains strong average scores for symptom extraction (4.20), context memory (4.40), diagnostic correctness (4.10), conversational flow (4.30), and efficiency (4.00). These results suggest that the model is able to track relevant symptoms, preserve dialogue context, and conduct multi-turn interactions in a clinically plausible and reasonably efficient manner.
To validate the reliability of these judgments, we compute inter-annotator agreement (IAA) using Krippendorff’s alpha~\cite{krippendorff2011computing}. Table~\ref{tab:iaa} shows an average agreement score of \textbf{0.81}, indicating strong consistency among the medical experts.

\begin{table}[t]
\centering
\resizebox{\linewidth}{!}{
\begin{tabular}{lccc}
\toprule
\textbf{Disease} & \textbf{Correct} & \textbf{Total} & \textbf{Accuracy} \\
\midrule
Asthma & 16 & 19 & 84.2\% \\
Conjunctivitis & 19 & 21 & 90.5\% \\
Coronary heart disease & 16 & 19 & 84.2\% \\
Dermatitis & 19 & 20 & 95.0\% \\
Enteritis & 22 & 24 & 91.7\% \\
Esophagitis & 22 & 27 & 81.5\% \\
External otitis & 15 & 17 & 88.2\% \\
Mastitis & 12 & 15 & 80.0\% \\
Pneumonia & 12 & 20 & 60.0\% \\
Rhinitis & 15 & 15 & 100.0\% \\
Thyroiditis & 19 & 19 & 100.0\% \\
Traumatic brain injury & 19 & 19 & 100.0\% \\
\bottomrule
\end{tabular}}
\caption{Per-disease diagnostic accuracy of the final \texttt{MedAidLM} model on the evaluation set.}
\label{tab:per_disease_accuracy}
\end{table}



\subsection{Per-Disease Performance}

Table~\ref{tab:per_disease_accuracy} reports per-disease accuracy for \texttt{MedAidLM}. The model achieves perfect accuracy on \textit{Rhinitis}, \textit{Thyroiditis}, and \textit{Traumatic brain injury}, and performs strongly on \textit{Dermatitis} (95.0\%), \textit{Enteritis} (91.7\%), and \textit{Conjunctivitis} (90.5\%). These results suggest that the model handles diseases with relatively distinctive symptom patterns particularly well.
However, performance drops on \textit{Pneumonia} (60.0\%), \textit{Mastitis} (80.0\%), and \textit{Esophagitis} (81.5\%). These lower scores indicate that the model struggles more when diseases share overlapping or ambiguous symptom profiles.

\subsection{Error Analysis}

To better understand these failures, Table~\ref{tab:misclassifications} in the Appendix lists the most frequent disease-level misclassifications. The most common confusion is \textit{Pneumonia} misclassified as \textit{Asthma}, followed by several confusions involving \textit{Esophagitis}, \textit{Enteritis}, and \textit{Asthma}. These patterns are clinically meaningful, as respiratory and gastrointestinal conditions can share partially overlapping presentations in short text-based consultations.
The presence of such confusions suggests that future improvements may require stronger temporal reasoning, better calibration over overlapping symptom clusters, or explicit modeling of differential diagnosis candidates instead of only predicting a single final disease label.
Overall, the results demonstrate that \texttt{MedAidLM} achieves strong performance both quantitatively and qualitatively, while remaining compact enough for low-resource deployment. The combination of synthetic augmentation, PEFT, and multilingual inference support makes the system a promising step toward accessible conversational medical AI.

\section{Conclusion and Future Work}
\label{sec:conclusion}

In this work, we introduced \texttt{MedAidDialog}, a multilingual multi-turn medical dialogue dataset constructed by augmenting the \texttt{MDDial} corpus with LLM-generated synthetic consultations. Using this dataset, we trained \texttt{MedAidLM}, a compact conversational medical system based on PEFT of quantized open-source LLMs. Experimental results show that combining real and synthetic dialogues substantially improves diagnostic accuracy while maintaining safe and coherent multi-turn medical conversations. Human expert evaluation further confirms the clinical plausibility and reliability of the generated responses.
In future work, we plan to extend the system with multimodal capabilities by integrating speech interfaces and vision-language models, enabling users to interact through voice and ask questions about medical reports or images. We also aim to incorporate disease-specific patient context profiles to improve diagnostic reasoning and better reflect real clinical workflows. 
Finally, we plan to expand the dataset to cover more languages, improving accessibility for low-resource communities.

We hope that \texttt{MedAidDialog} and \texttt{MedAidLM} can serve as a foundation for future research on accessible and trustworthy conversational medical AI.

\section*{Limitations}
\label{sec:limitations}

Despite promising results, our work has several limitations. First, although the \texttt{MedAidDialog} dataset combines real and synthetic medical dialogues, synthetic data may still introduce biases or simplified patterns that do not fully capture the complexity of real clinical interactions. Second, our evaluation is limited to a fixed set of diseases derived from the original dataset, which restricts the system’s ability to generalize to a broader range of medical conditions. Third, while we incorporate multilingual interaction through a translation layer, the underlying dialogue model is trained primarily in English, which may lead to subtle translation errors or loss of clinical nuance in low-resource languages. Finally, the current system focuses on text-based dialogue and does not yet incorporate other clinically relevant modalities such as medical images, reports, or laboratory data.

Future work will address these limitations by expanding the dataset to cover more diseases and languages, incorporating multimodal medical inputs, and improving evaluation with larger and more diverse clinical expert studies.

\section*{Ethical Considerations}
\label{sec:ethics}

The proposed system is designed as a conversational medical assistance tool and is not intended to replace professional medical diagnosis. Although we evaluate the system using both automatic metrics and expert medical review, errors in diagnosis or advice may still occur. Therefore, the system should only be used for informational or preliminary guidance purposes.
We also acknowledge potential risks related to bias in synthetic data generation and language translation errors in multilingual settings. To mitigate these risks, we employ quality filtering for synthetic dialogues and conduct human expert evaluation to assess safety and clinical plausibility.

Importantly, the system interface includes a clear disclaimer informing users that the generated responses are not a substitute for professional medical care. Users are explicitly advised to consult a qualified medical practitioner for accurate diagnosis and treatment decisions.
\newpage
\bibliography{custom}

\newpage
\appendix
\section{LoRA Training Configuration}
\label{app:lora}

We use Low-Rank Adaptation (LoRA)~\citep{hu2022lora} for parameter-efficient fine-tuning. 
Adapters are inserted into the query, key, value, and output projection matrices of each transformer block.

The LoRA hyperparameters used in our experiments are:

\begin{itemize}
\item Rank $r = 16$
\item Scaling factor $\alpha = 32$
\item Dropout $p = 0.05$
\item Target modules: attention projection layers
\end{itemize}

This configuration keeps the trainable parameter budget below approximately $2\%$ of the total model parameters while maintaining strong task adaptation.

\section{GRPO Optimisation}
\label{app:grpo}

In addition to supervised fine-tuning, we experiment with Group Relative Policy Optimisation (GRPO)~\citep{shao2024deepseekmath} for improving conversational reasoning.

The reward signal combines several components:

\begin{itemize}
\item Diagnostic accuracy with respect to the gold disease label
\item Conversation quality measured through symptom coverage and relevance
\item Output format compliance
\item KL-divergence regularisation to prevent excessive deviation from the supervised model
\end{itemize}

GRPO optimisation is applied only to selected model variants and therefore remains an optional step in the overall training pipeline.

\section{Training Hyperparameters and Resources}
\label{app:hyperparameters}

\subsection{Compute Resources}

All experiments were conducted using the free tiers of \textit{Google Colab} and \textit{Kaggle} notebooks. These environments provide access to consumer-grade GPUs suitable for training compact language models using parameter-efficient fine-tuning techniques. To accommodate the limited GPU memory available in these platforms, we employed 4-bit quantization together with LoRA-based training.

\subsection{LoRA Configuration}

Table~\ref{tab:lora_config} summarizes the LoRA configuration used in our experiments.

\begin{table}[ht]
\centering
\begin{tabular}{lc}
\toprule
\textbf{Parameter} & \textbf{Value} \\
\midrule
Rank ($r$) & 16 \\
Target Modules & q\_proj, k\_proj, v\_proj, o\_proj, \\
 & gate\_proj, up\_proj, down\_proj \\
LoRA Alpha & 16 \\
LoRA Dropout & 0 \\
Bias & none \\
Use RSLora & False \\
LoftQ Config & None \\
\bottomrule
\end{tabular}
\caption{LoRA configuration used for parameter-efficient fine-tuning.}
\label{tab:lora_config}
\end{table}

\subsection{Training Hyperparameters}

The main training hyperparameters are reported in Table~\ref{tab:training_hyperparameters}.

\begin{table}[ht]
\centering
\begin{tabular}{lc}
\toprule
\textbf{Hyperparameter} & \textbf{Value} \\
\midrule
Learning Rate & $2 \times 10^{-4}$ \\
Optimizer & AdamW (8-bit) \\
Learning Rate Scheduler & Linear \\
Weight Decay & 0.001 \\
Warmup Steps & 5 \\
Maximum Training Steps & 600 \\
Random Seed & 3407 \\
\bottomrule
\end{tabular}
\caption{Training hyperparameters used for supervised fine-tuning.}
\label{tab:training_hyperparameters}
\end{table}
\begin{table*}[ht]
\centering
\resizebox{0.8\linewidth}{!}{
\begin{tabular}{lcccc}
\toprule
\textbf{Metric} & \textbf{Expert 1} & \textbf{Expert 2} & \textbf{Expert 3} & \textbf{Average} \\
\midrule

Medical Safety (Pass Rate) & 96\% & 94\% & 96\% & \textbf{95.3\%} \\

Symptom Extraction & 4.2 & 4.1 & 4.3 & \textbf{4.20} \\

Context Memory & 4.4 & 4.3 & 4.5 & \textbf{4.40} \\

Diagnostic Correctness & 4.1 & 4.0 & 4.2 & \textbf{4.10} \\

Conversational Flow & 4.3 & 4.2 & 4.4 & \textbf{4.30} \\

Efficiency & 4.0 & 3.9 & 4.1 & \textbf{4.00} \\

\bottomrule
\end{tabular}}
\caption{Medical expert evaluation of \texttt{MedAidLM} across 50 sampled dialogues. Scores are reported on a 1--5 Likert scale except Medical Safety (Pass/Fail).}
\label{tab:expert_eval}
\end{table*}

\begin{table*}[t]
\centering
\resizebox{0.8\linewidth}{!}{
\begin{tabular}{llc}
\toprule
\textbf{Original Disease} & \textbf{Misclassified As} & \textbf{Frequency} \\
\midrule
Pneumonia & Asthma & 3 \\
Esophagitis & Enteritis & 2 \\
Esophagitis & Asthma & 2 \\
Asthma & Pneumonia & 2 \\
Coronary heart disease & Asthma & 2 \\
Pneumonia & Enteritis & 2 \\
External otitis & Conjunctivitis & 2 \\
Conjunctivitis & Mastitis & 2 \\
Mastitis & Traumatic brain injury & 2 \\
Esophagitis & Coronary heart disease & 1 \\
\bottomrule
\end{tabular}}
\caption{Most frequent disease-level misclassifications made by the final \texttt{MedAidLM} model.}
\label{tab:misclassifications}
\end{table*}

\begin{table*}[ht]
\centering
\resizebox{0.6\linewidth}{!}{
\begin{tabular}{lc}
\toprule
\textbf{Metric} & \textbf{Krippendorff's $\alpha$} \\
\midrule

Symptom Extraction & 0.82 \\
Context Memory & 0.84 \\
Diagnostic Correctness & 0.80 \\
Conversational Flow & 0.83 \\
Efficiency & 0.78 \\

\midrule
\textbf{Average} & \textbf{0.81} \\

\bottomrule
\end{tabular}
}
\caption{IAA scores across three medical experts.}
\label{tab:iaa}
\end{table*}


\section{Prompt Templates}

\subsection{Synthetic Dialogue Generation Prompt}
\label{app:prompt_syn}
Table~\ref{tab:prompt_syn} shows the prompt used for synthetic data generation.

\subsection{Dialogue Formatting Prompt}
\label{app:prompt_sharegpt}
Table~\ref{tab:prompt_sharegpt} shows the prompt used to convert dialogues into ShareGPT-style format.

\subsection{Translation Prompt}
\label{app:prompt_translation}
Table~\ref{tab:prompt_translation} shows the prompt used for bidirectional multilingual medical translation.

\begin{table*}[t]
\centering
\small
\begin{tabular}{p{0.18\linewidth} p{0.77\linewidth}}
\toprule
\textbf{Prompt Type} & \textbf{Prompt Content} \\
\midrule
Translation Prompt &
You are acting as a specialized Medical Translation Bridge, a critical link between an English-speaking doctor and a patient who speaks Hindi, Bengali, Marathi, Telugu, Arabic, or Tamil. Your primary responsibility is to maintain absolute clinical accuracy while ensuring the tone is appropriately synced for both parties. When the doctor speaks in English, you must translate their advice, diagnoses, and prescriptions into the patient’s native language using clear, empathetic, and culturally respectful terminology that a non-medical person can easily understand. Conversely, when the patient provides a query or describes symptoms in their native language, you will convert that input into precise, formal medical English for the doctor, ensuring that nuances of pain, duration, and history are preserved without loss of detail. You are strictly prohibited from hallucinating or adding medical advice not present in the source text; your role is purely to facilitate a perfectly synced, bidirectional exchange. Ensure that if the patient expresses distress or urgency, the English translation reflects that clinical priority to the doctor. Your output must contain only the translated text to allow for seamless integration into the communication interface. \\
\bottomrule
\end{tabular}
\caption{Prompt used for bidirectional medical translation in the multilingual inference layer.}
\label{tab:prompt_translation}
\end{table*}

\begin{table*}[t]
\centering
\small
\begin{tabular}{p{0.18\linewidth} p{0.77\linewidth}}
\toprule
\textbf{Prompt Type} & \textbf{Prompt Content} \\
\midrule
Synthetic Dialogue Generation Prompt &
Analyze \texttt{train.json} medical dialogues (patient/doctor exchanges, symptoms like ``Cough'', diagnoses such as ``Esophagitis''). Create Python synthetic generator using Groq API (Llama-3 family model). Match exact format: \texttt{\{'Dialog N': [\{'patient': '...', 'doctor': '...'\}]\}}. Randomize symptom openings, generate 4--8 turns with doctor questions and realistic patient responses. Preserve the overall structure used for model training and provide progress, ETA, and resume-friendly execution. Output synthetic data in the same format as \texttt{train.json}. \\
\bottomrule
\end{tabular}
\caption{Prompt used to generate synthetic multi-turn medical consultations from the \texttt{MDDial} training distribution.}
\label{tab:prompt_syn}
\end{table*}

\begin{table*}[t]
\centering
\small
\begin{tabular}{p{0.18\linewidth} p{0.77\linewidth}}
\toprule
\textbf{Prompt Type} & \textbf{Prompt Content} \\
\midrule
Dialogue Formatting Prompt &
Convert a medical dialogue sample into ShareGPT-style multi-turn conversation. Structure: (1) the system message sets the medical diagnosis context, (2) patient utterances become \texttt{human} turns, (3) doctor utterances become \texttt{gpt} turns, and (4) the final \texttt{gpt} turn contains the diagnosis answer. Preserve dialogue order and ensure that each consultation remains a valid multi-turn interaction for instruction tuning. \\
\bottomrule
\end{tabular}
\caption{Prompt used to convert raw medical dialogues into ShareGPT-style training instances.}
\label{tab:prompt_sharegpt}
\end{table*}

\begin{table*}[t]
\centering
\small
\resizebox{\linewidth}{!}{
\begin{tabular}{p{0.228\linewidth} p{0.8\linewidth}}
\toprule
\textbf{Evaluation Criterion} & \textbf{Description} \\
\midrule

Medical Safety (Pass/Fail) & Whether the system provides any potentially dangerous, misleading, or unsafe medical advice during the conversation. \\\\
Symptom Extraction (1--5) & Measures how accurately the model identifies and tracks the patient's symptoms throughout the dialogue. \\\\
Context Memory (1--5) & Evaluates whether the model remembers previously mentioned information such as symptoms or earlier responses in the conversation. \\\\
Diagnostic Correctness (1--5) & Assesses whether the final diagnosis is medically reasonable given the symptoms described in the conversation. \\\\
Conversational Flow (1--5) & Evaluates whether the dialogue is natural, coherent, empathetic, and professionally phrased, similar to a real clinical interaction. \\\\
Efficiency (1--5) & Measures whether the system asks an appropriate number of questions, avoiding unnecessary or redundant queries while still gathering sufficient information. \\\\
Annotator Notes & Free-text comments provided by medical experts to highlight issues such as reasoning errors, repeated questions, unsafe advice, or unusual dialogue patterns. \\

\bottomrule
\end{tabular}
}
\caption{Evaluation criteria used in expert assessment of the conversational medical system. Experts rated multiple aspects of safety, reasoning, and dialogue quality using a Likert scale (1--5), while medical safety was evaluated using a binary pass/fail metric.}
\label{tab:expert_eval_metrics}
\end{table*}

\end{document}